\documentclass{article}
\usepackage{ijcai19}

\usepackage{times}
\usepackage{soul}
\usepackage{url}
\usepackage[utf8]{inputenc}
\usepackage{graphicx}
\usepackage{amsmath}
\usepackage{booktabs}
\usepackage{algorithm}
\usepackage{color}
\usepackage{chngcntr}
\usepackage[noend]{algpseudocode}
\usepackage{listings}
\usepackage{courier}
\usepackage{soul}
\usepackage{amsmath}
\usepackage{mathtools}
\usepackage{multirow}
\usepackage{graphicx}
\usepackage{soul}
\usepackage[flushleft]{threeparttable}
\usepackage{amssymb} 
\usepackage{array}
\usepackage{subcaption}

\usepackage{tikz}
\usetikzlibrary{positioning,shapes,chains}

\usepackage{makecell}

\usepackage{booktabs}

\usepackage{caption}
\usepackage{amsthm}
\usepackage{thmtools}
\declaretheorem[style=definition,qed={\small$\blacksquare$}]{definition}
\declaretheorem[style=definition]{proposition}

\usepackage{flushend}

\DeclareMathSymbol{\Gamma}{\mathord}{operators}{"00}
\DeclareMathSymbol{\Delta}{\mathord}{operators}{"01}
\DeclareMathSymbol{\Theta}{\mathord}{operators}{"02}
\DeclareMathSymbol{\Lambda}{\mathord}{operators}{"03}
\DeclareMathSymbol{\Xi}{\mathord}{operators}{"04}
\DeclareMathSymbol{\Pi}{\mathord}{operators}{"05}
\DeclareMathSymbol{\Sigma}{\mathord}{operators}{"06}
\DeclareMathSymbol{\Upsilon}{\mathord}{operators}{"07}
\DeclareMathSymbol{\Phi}{\mathord}{operators}{"08}
\DeclareMathSymbol{\Psi}{\mathord}{operators}{"09}
\DeclareMathSymbol{\Omega}{\mathord}{operators}{"0A}

\DeclareFontFamily{OT1}{pzc}{}
\DeclareFontShape{OT1}{pzc}{m}{it}{<-> s * [1.20] pzcmi7t}{}
\DeclareMathAlphabet{\mathpzc}{OT1}{pzc}{m}{it}

\DeclareFontFamily{U}{mathc}{}
\DeclareFontShape{U}{mathc}{m}{it}{<->s*[1.03] mathc10}{}
\DeclareMathAlphabet{\mathscr}{U}{mathc}{m}{it}

\DeclareMathAlphabet{\mathcalligra}{T1}{calligra}{m}{n}

\DeclareMathAlphabet{\mathsfit}{\encodingdefault}{\sfdefault}{m}{sl}

\newcommand{\mc}[1]{\mathcal{#1}}

\newcommand{\mcr}[1]{\mathscr{#1}}

\definecolor{red}{rgb}{1.00,0.00,0.00}
\definecolor{blue}{rgb}{0.00,0.00,1.00}
\definecolor{green}{rgb}{0.4,1.00,0.0}
\definecolor{yellow}{rgb}{0.5,0.5,0.0}
\definecolor{gray}{rgb}{0.5,0.5,0.5}
\definecolor{ipython_frame}{RGB}{207, 207, 207}
\definecolor{halfgray}{gray}{0.3}

\algrenewcommand\alglinenumber[1]{\tiny\color{black} #1~}

\lstdefinestyle{customlst}{
    mathescape,
    aboveskip=3pt,
    belowskip=-15pt,
    numbersep=2pt,
    numberstyle=\tiny\color{halfgray},
    captionpos=b,
    stepnumber=1,
    basicstyle=\ttfamily\fontsize{6.6}{0}\selectfont,
    keywordstyle=\bfseries}
\makeatletter
\def\moverlay{\mathpalette\mov@rlay}
\def\mov@rlay#1#2{\leavevmode\vtop{%
   \baselineskip\z@skip \lineskiplimit-\maxdimen
   \ialign{\hfil$\m@th#1##$\hfil\cr#2\crcr}}}
\newcommand{\charfusion}[3][\mathord]{
    #1{\ifx#1\mathop\vphantom{#2}\fi
        \mathpalette\mov@rlay{#2\cr#3}
      }
    \ifx#1\mathop\expandafter\displaylimits\fi}
\makeatother

\newcommand{\figref}[1]{Fig.~\ref{fig:#1}}

\newcommand{\lstref}[1]{Listing~\ref{lst:#1}}

\newcommand{\Half}{\frac{1}{2}}

\newcommand{\Init}{\fontfamily{cmss}\selectfont\text{init}}
\newcommand{\Static}{\fontfamily{cmss}\selectfont\text{static}}
\newcommand{\End}{\fontfamily{cmss}\selectfont\text{end}}
\newcommand{\Struc}{\textit{Struc}}

\newcommand{\kjoin}{\sqcup}
\newcommand{\KJoin}{\bigsqcup}
\newcommand{\Voc}{\mcr{V}}

\usepackage[notext,nott,frenchstyle,partialup,notextcomp]{kpfonts}
\usepackage[mathcal,mathscr]{euscript}

\usepackage{eucal}
\usepackage[frenchstyle,expert]{mathdesign}
\usepackage{eulervm}

\usepackage{xspace}

\newcommand{\StackDom}{\textsc{stack}\xspace}
\newcommand{\Rover}{\textsc{rover}\xspace}
\newcommand{\Can}{{\fontfamily{cmss}\selectfont\text{canon}}}

\newcommand{\sans}[1]{{\fontfamily{cmss}\selectfont{#1}}}

\newcommand{\CM}[1]{{\fontfamily{cmss}\selectfont#1}\xspace}

\title{Computing the Scope of Applicability for Acquired 
Task Knowledge\\in Experience-Based Planning Domains}

\author{
Vahid Mokhtari$^1$
\and
Lu\'{i}s Seabra Lopes$^1$\and
Armando J. Pinho$^{1}$\And
Roman Manevich$^2$
\affiliations
$^1$The University of Aveiro, Portugal\\
$^2$The University of Texas at Austin, USA
\emails\normalsize
\{mokhtari.vahid, lsl, ap\}@ua.pt,
romanm@cs.bgu.ac.il
}

\begin{document}

\maketitle

\begin{abstract}\small
Experience-based planning domains have been proposed to improve 
problem solving by learning from experience. 
They rely on acquiring and using task knowledge, i.e., activity 
schemata, for generating solutions to problem instances in a class 
of tasks.
Using Three-Valued Logic Analysis (TVLA), we extend previous work to 
generate a set of conditions that determine the \emph{scope of 
applicability} of an activity schema.
The inferred scope is a bounded representation of a set of problems 
of potentially unbounded size, in the form of a 3-valued logical 
structure, which is used to automatically find an applicable activity 
schema for solving task problems. We validate this work in two classical 
planning domains.
\end{abstract}

\section{Introduction}\label{sec:introduction}

Planning is a key ability for intelligent robots, increasing their autonomy and 
flexibility through the construction of sequences of actions to achieve their goals 
\cite{ghallab2004automated}.
Planning is a hard problem and even what is known historically as 
classical planning is PSPACE-complete over propositional state variables 
\cite{bylander1994computational}.
To carry out increasingly complex tasks, robotic communities make strong 
efforts on developing robust and sophisticated high-level decision making 
models and implement them as planning systems. 
One of the most challenging issues is to find an optimum in a trade-off 
between computational efficiency and needed domain expert engineering 
work to build a reasoning system. In a recent work, Mokhtari \emph{et al}. 
[\citeyear{mokhtari2016jint,mokhtari2016icaps,vahid2017prletter,vahid2017iros}]
have proposed and integrated the notion of \emph{Experience-Based Planning 
Domain} (EBPD)---a framework that integrates important concepts for 
long-term learning and planning---into robotics. 
An EBPD is an extension of the standard \emph{planning domains} which 
in addition to planning operators, includes experiences and methods 
(called \emph{activity schemata}) for solving classes of problems. 
The EBPDs framework consist of three components: \emph{experience 
extraction}, \emph{conceptualization} and \emph{planning}. 
Experience extraction provides a human-robot interaction for teaching 
tasks and recording experiences of past robot's observations 
and activities.
Experiences are used to learn activity schemata, i.e., methods of guiding a 
search-based planner for finding solutions to other related problems. 
Conceptualization combines several techniques, including deductive 
generalization, different forms of abstraction, feature extraction and loop 
detection to generate activity schemata 
from experiences. 
Planning is a hierarchical problem solver 
which applies learned activity schemata 
for problem solving. 
In previous work, algorithms have been developed for experience extraction, 
activity schema learning and task planning 
[Mokhtari \emph{et al}. \citeyear{mokhtari2016jint,vahid2017prletter,vahid2017iros}].

As a contribution of this paper, 
we extend and improve the EBPDs framework to automatically retrieve an 
applicable activity schema for solving a task problem. 
We propose an approach to infer a set of conditions from an experience 
that determines the \emph{scope of applicability} of an activity schema 
for solving a set of task problems. 
The inferred scope is a $3$-valued logical structure \cite{kleene1952introduction} 
(i.e., a structure that extends Boolean logic by introducing an indefinite 
value $\Half$ to denote either $0$ or $1$) which associates a bounded 
representation for a set of $2$-valued logical structures of potentially 
unbounded size. 
We employ Three-Valued Logic Analysis (TVLA) \cite{SRW:TOPLAS02} 
both to infer the scope of applicability of activity schemata and to test whether 
existing activity schemata can be used to solve given task problems. 

We recapitulate the prior work and present our approach to abstracting 
an experience and inferring the scope of applicability of an activity schema 
using the TVLA. 
We validate our system over two classical planning domains.

\section{Related Work}\label{sec:literature}

The EBPDs' objective is to perform tasks.
Learning of Hierarchical Task Networks (HTNs) is among the most
related works to EBPDs.
In HTN planning, a plan is generated by decomposing a method for a given
task into simpler tasks until primitive tasks are reached that can be
directly achieved by planning operators.
CaMeL \cite{ilghami2002camel,ilghami2005learning} is an HTN learner which 
receives as input plan traces and the structure of an HTN method and tries 
to identify under which conditions the HTN is applicable. 
CaMeL requires all information about methods except for the preconditions.
The same group transcends this limitation in a later work \cite{ilghami2006hdl}
and presents the HDL algorithm which starts with no
prior information about the methods but requires hierarchical plan
traces produced by an expert problem-solver.
HTN-Maker \cite{hogg2008htn,hogg2016learning} generates an HTN domain model 
from a STRIPS 
domain model, a set of STRIPS plans, and a set of annotated tasks.
HTN-Maker generates and traverses a list of states by applying the actions
in a plan, and looks for an annotated task whose effects and preconditions
match some states. Then it regresses the effects of the annotated task
through a previously learned method or a new primitive task.
Overall, identifying the hierarchical structure is an issue,
and most of the techniques in HTN learning rely on the hierarchical
structure of the HTN methods specified by a human expert.
By contrast, the EBPDs framework presents a fully autonomous
approach to learning activity schemata with loops (an alternative 
to recursive HTN methods) from single experiences.

Aranda \cite{Srivastava2011615} takes a planning problem and finds 
a plan that includes loops. 
Using TVLA~\cite{LAmiS:SAS00} and back-propagation, Aranda finds 
an abstract state space from a set of concrete states of problem 
instances with varying numbers of objects
that guarantees completeness, i.e., the plan works for all
inputs that map onto the abstract state.
These strong guarantees come at a cost: (i) restrictions on the language
of actions; and (ii) high running times. Indeed computing the abstract state 
is worst-case doubly-exponential in the number of predicates.
In contrast, the EBPDs system assumes standard PDDL actions. We also use 
TVLA to compute an abstract structure that determines 
the scope of applicability of an activity 
schema, however, we trade completeness for a polynomial time algorithm, 
which results in dramatically better performance.

Loop\textsc{Distill} \cite{Winner07loopdistill} also 
learns plans with loops from example plans.
It identifies the largest matching sub-plan in a given example
and converts the repeating occurrences of the sub-plans into
a loop. The result is a domain-specific planning program
(dsPlanner), i.e., a plan with if-statements and while-loops
that can solve similar problems of the same class.
Loop\textsc{Distill}, nonetheless, does not address the applicability test of plans.

Other approaches in AI planning including
case based planning \cite{hammond1986chef,borrajo2015acm},
and macro operators \cite{fikes1972strips2,chrpa2010generation}
can also be related to our work.
These methods tend to suffer from the utility problem, in which
learning more information can be counterproductive due to
the difficulty with storage and management of the information
and with determining which information should be used to solve a
particular problem.
In EBPDs, by combining generalization with abstraction in task learning,
it is possible to avoid saving large sets of concrete cases.
Additionally, since in EBPDs, task learning is supervised,
solving the utility problem can be to some extent delegated to
the user, who chooses which tasks and associated procedures to teach.

\section{The Prior Work}\label{sec:prior}

An EBPD $\mc{D=\langle A,O,E,M \rangle}$ relies on a set of abstract 
planning operators $\mc A$, a set of concrete planning operators $\mc O$, 
a set of experiences $\mc E$, and a set of activity schemata (i.e., task 
planning models) $\mc M$, for problem solving. 

Any planning operator $o\in(\mcr{A\lor O})$ is a 
tuple $\langle h,S,P,E \rangle$, where $h$ is the operator head, 
$S$ is a set of atoms describing the static part of the world 
information, $P$ is the precondition (a conjunction of atoms that must be 
available in a state in order to apply $o$), and $E$ is the effect (a set 
of atoms that specifies the changes on a state effected by $o$). 
The abstract and concrete planning operators are linked together using 
an operator abstraction hierarchy (specified by a \sans{parent} property in 
concrete planning operators), 
for example, 
\sans{(pick ?block ?table)}
is an abstract operator for the concrete operator 
\sans{(pick ?hoist ?block ?table ?location)}. 
More in \cite{vahid2017prletter}. 

An \emph{experience} $e \in\mc{E}$ is a triple of ground 
structures $\langle t,K,\pi \rangle$, where $t$ is the task achieved 
in an experience, e.g., \sans{{(stack t1 t2)}}, $K$ is a set of 
key-properties, i.e., a set of predicates with temporal symbols, 
to describe the experience, and $\pi$ is a plan to achieve $t$. 
The temporal symbols for representing key-properties are:
{\Static}---always true during an experience,
{\Init}---true at the initial state,
and {\End}---true at the final state.
\lstref{experience} shows part of an experience 
\footnote{\fontsize{8pt}{8pt}\selectfont 
An approach for teaching a robot to achieve a task 
and extracting experiences has been presented in 
[Mokhtari \emph{et al}. \citeyear{vahid2014experience,mokhtari2016jint}]. 
}.
This experience is used to illustrate the proposed approach in this paper.

\begin{lstlisting}[style=customlst,
    float=!t,
    label=lst:experience,
    % abovecaptionskip=10pt,
    morekeywords={define, experience, domain, episode_id,
    task, parameters, key, properties, plan, objects},
    caption={Part of the `stack' experience in the \StackDom domain. 
    There are 8 (4 blue and 4 red) blocks in this experience. The goal 
    of the task in this experience is to stack the blocks (that are 
    initially on a table) on a pile with blue blocks at the bottom and 
    red blocks on the top. 
    The key-properties describe the initial, final and static world 
    information of the experience. The solution plan contains 31 actions. 
    % (some information is omitted due to limited space).
    }]
(:task stack
 :parameters     (t1 t2)
 :key-properties ((static(table t1))
                  (static(pile t2))
                  (static(location l1))
                  (static(hoist h1))
                  (static(attached t2 l1))
                  (static(attached t1 l1))
                  (static(belong h1 l1))
                  (static(pallet p1))
                  (static(block b1))
                  (static(block b2))
                  $\cdots$
                  (static(block b7))
                  (static(block b8))
                  (static(blue b1))
                  (static(blue b2))
                  $\cdots$
                  (static(red b7))
                  (static(red b8))
                  (init(top p1 t2))
                  (init(ontable b1 t1))
                  (init(ontable b2 t1))
                  $\cdots$
                  (init(ontable b7 t1))
                  (init(ontable b8 t1))
                  (init(at h1 t1))
                  (init(empty h1))
                  (end(on b1 p1))
                  (end(on b2 b1))
                  $\cdots$
                  (end(on b7 b6))
                  (end(on b8 b7))
                  (end(top b8 t2)))
 :plan           ((pick h1 b1 t1 l1)
                  (move h1 t1 t2 l1)
                  (stack h1 b1 p1 t2 l1)
                  (move h1 t2 t1 l1)
                  (pick h1 b2 t1 l1)
                  (move h1 t1 t2 l1)
                  (stack h1 b2 b1 t2 l1)
                  $\cdots$
                  (pick h1 b8 t1 l1)
                  (move h1 t1 t2 l1)
                  (stack h1 b8 b7 t2 l1)))
\end{lstlisting}

An \emph{activity schema} $m \in\mc{M}$ 
is a pair of ungrounded structures $\langle h,\Omega \rangle$, where $h$ is 
the target task, e.g., \sans{(stack ?t1 ?t2)},
and $\Omega$ is an abstract plan to achieve the task in $m$, i.e., a sequence 
or loops of abstract operators. Each abstract operator in the abstract plan is 
in the form $(a,F)$ where $a$ is an abstract operator head, and $F$ is a set 
of features, i.e., key-properties in an experience that describe the 
arguments of $a$. 
A concrete example of an activity schema is given in the rest of the paper.

\subsection{Acquiring Activity Schemata in EBPDs}\label{sec:schema}

Activity schemata are acquired from single experiences through 
a conceptualization methodology: 

Following the tradition of PLANEX \cite{fikes1972strips2} and 
Explanation-Based Generalization [Mitchell \emph{et al}.,~\citeyear{mitchell1986explanation}], 
all constants appearing in the actions as well as 
in the key-properties of an experience are variablized. The obtained 
generalized experience forms the basis of the activity schema.

After the generalization, concrete actions in the plan of the generalized 
experience are replaced with abstract actions, as specified in the operator 
abstraction hierarchy. 
That is, some concrete actions are excluded from the abstract plan, and 
some arguments of the concrete actions are excluded from the arguments 
of the respective abstract actions. 

Then, all potential key-properties 
(i.e., features) that link the arguments of the abstract actions with the 
parameters of the experience are extracted and associated to the abstract 
actions. For example in \lstref{experience}, the key-property 
\sans{(init(ontable b1 t1))} is a feature that links \sans{b1}, 
an argument of an (abstract) action \sans{pick}, to \sans{t1}, 
a parameter of the task \sans{stack}. 
During problem solving features determine which objects in a given problem 
are preferable to instantiate abstract actions. 

Finally, potential loops of actions in the activity schema are detected.
Mokhtari \emph{et al}. \shortcite{vahid2017prletter} propose 
a {Contiguous Non-overlapping Longest Common Prefix} (CNLCP) algorithm. 
CNLCP is an extension of the standard function~of constructing the 
Longest-Common-Prefix (LCP) array---an array storing the lengths 
of the longest common-prefixes of consecutive suffixes in a 
suffix array \cite{manber1993suffix}.
CNLCP first computes a {Non-overlapping} LCP (NLCP) array between 
all consecutive suffixes (in a suffix array) such that the lengths 
of the longest common prefixes between every two suffixes must be at 
most equal to the difference in lengths between the two suffixes, and 
then preserves only consecutive NLCPs.
When a loop is detected, the respective loop iterations are merged 
together. \lstref{schema_loop} shows part of the learned activity 
schema for the `stack' experience in Listing~\ref{lst:experience}. 

\subsection{Task Planning in EBPDs}
\label{sec:planner}

Task planning in EBPDs is achieved by a hierarchical planning system, called 
\textit{Schema-Based Planner} (SBP), consisting of an abstract and a concrete 
planner \cite{vahid2017prletter}. 
A \emph{task planning problem} is a tuple 
$\mc P=\langle t,\sigma,s_0,g \rangle$ where $t$ is the target task, 
e.g., \sans{(stack t1 t2)}, 
$\sigma$ is static world 
information, $s_0$ is the initial state, and $g$ is the goal.

Given the abstract and concrete planning operators $(\mcr{A,O})$, 
and an activity schema $m$, SBP applies $m$ to generate a plan for $\mcr{P}$. 
The abstract planner first drives an abstract solution to $\mcr{P}$ by 
generating instances of the abstract actions (of the abstract plan) in $m$. 
It also extends possible loops in $m$ for the applicable objects in $\mcr{P}$. 
To extend a loop, the abstract planner simultaneously generates all successors 
for an iteration of the loop as well as for the following abstract action after 
the loop. 
It then computes a cost for all generated successors based on the number of 
features of abstract actions (in $m$) verified with the features extracted for 
the instantiated abstract actions, and selects the best current action with the 
lowest cost during the search. 
Finally, the abstract planner generates a ground abstract plan when it gets 
the end of (the abstract plan of) $m$. 

The produced ground abstract plan becomes the main skeleton of the final 
solution based on which the concrete planner generates a final plan by 
instantiating and substituting concrete actions for the abstract actions 
(as specified in the operator abstraction hierarchy). 

See [Mokhtari \emph{et al}. \citeyear{vahid2017iros,vahid2017prletter}] 
for the algorithms of learning activity schemata 
and task planning in EBPDs. 

\begin{lstlisting}[style=customlst,
    label=lst:schema_loop,
    float=t,
    captionpos=b,
    % aboveskip=-5pt,
    % belowskip=-5pt,
    % abovecaptionskip=10pt,
    belowcaptionskip=7pt,
    morekeywords={parameters, domain, objects,
    define, activity, schema, method, abstract, plan, precondition}, 
    caption={Part of a learned activity schema for the `stack' task~with 
    two loops. 
    From Listing~\ref{lst:experience}, the constants are replaced with 
    \mbox{variables} (Generalization), some 
    actions are excluded from the abstract plan (Operator 
    abstraction), abstract actions are associated with features (Feature 
    extraction), and repetitive abstract actions with same features form 
    loops (Loop detection). 
    See \protect\cite{vahid2017prletter}.
    }]
(:method stack
 :parameters (?t1 ?t2)
 :abstract-plan
  ((!pick ?b1 ?t1)
      ((static(blue ?b1))$\cdots$)
   (!stack ?b1 ?p1 ?t2)
      ((static(blue ?b1))(static(pallet ?p1))$\cdots$)
   $\text{\textbf{(loop}}$ (!pick ?b2 ?t1)
            ((static(blue ?b2))$\cdots$)
         (!stack ?b2 ?b1 ?t2)
            ((static(blue ?b1))(static(blue ?b2))$\cdots$)$\text{\textbf{)}}$
   (!pick ?b5 ?t1)
      ((static(red ?b5))$\cdots$)
   (!stack ?b5 ?b4 ?t2)
      ((static(blue ?b4))(static(red ?b5))$\cdots$)
   $\text{\textbf{(loop}}$ (!pick ?b6 ?t1)
            ((static(red ?b6))$\cdots$)
         (!stack ?b6 ?b5 ?t2)
            ((static(red ?b5))(static(red ?b6))$\cdots$)$\text{\textbf{)}}$
   (!pick ?b8 ?t1)
      ((static(red ?b8))(end(top ?b8 ?t2))$\cdots$)
   (!stack ?b8 ?b7 ?t2)
      ((static(red ?b7))(static(red ?b8))$\cdots$)))
\end{lstlisting}

\section{Inferring the Scope of Applicability}
\label{sec:tvla_learning}

In the previous work, the EBPDs framework lacked a strategy to 
find an applicable activity schema, among several learned activity schemata, 
for solving a task problem.
We extend the EBPDs framework to infer the scope of an activity schema 
from the key-properties of an experience in 
the form of a \emph{$3$-valued logical structure}. 
This allows for the applicability test of an activity schema to solve a  
set of task problems. We employ Canonical Abstraction \cite{SRW:TOPLAS02} 
which associates a bounded representation for any (possibly 
infinite) set of logical structures of potentially unbounded size. 

To infer the scope of an activity schema, we first represent (the 
key-properties of) an experience in a $2$-valued structure: 

\begin{definition}
\label{def:2_valued}
A \emph{$2$-valued logical structure}, also called a \emph{concrete structure}, 
over a finite set of predicates $\Voc$ is a pair, 
$S=\langle U, \iota\rangle$, where $U$ is the universe of the $2$-valued 
structure 
and $\iota$ is the interpretation function that maps predicates to their 
truth-values in the structure: for every predicate $p^k\in \Voc$ of arity 
$k$, $\iota(p):U^k \to \{0,1\}$.
\end{definition}

We convert a set of key-properties $K$ to the $2$-valued 
structure $\Struc(K) =(U, \iota)$ as follows:
\vspace{-3pt}
\[\fontfamily{cmss}\selectfont\small
\begin{array}{rcl}
U     &=& \bigcup\limits_{\tau(p(t_1,\ldots,t_k)) \in K}\{t_1,\ldots,t_k\}\\
\Voc  &=& \bigcup\limits_{\tau(p(t_1,\ldots,t_k)) \in K}\big\{\tau(p)~|~\tau\in
          \{\small \Static, \Init, \End\}\big\}\\
\iota &=& \lambda \tau(p^k)\in\Voc. \\ 
      & & \lambda (t_1,\ldots,t_k)\in {U}^k.\ 
                    \left\{
                           \begin{array}{ll}
                             1, & \hbox{if}\enspace\tau(p(t_1,\ldots,t_k)) \in K\hbox{;} \\
                             0, & \hbox{otherwise.}
                           \end{array}
                    \right.
\end{array}
\]
\vspace{-5pt}

That is, the universe of $\Struc(K)$ consists of the objects appearing in the 
key-properties of $K$, and the interpretation is defined over the key-properties 
of $K$.
The interpretation of a key-property $\tau(p)$, where $\tau \in \{\Static, \Init, \End\}$, 
is $1$ if the corresponding key-property appears in $K$; and $0$ otherwise.

\begin{figure}[!t]
    \centering
    \begin{subfigure}[b]{\linewidth}
        \includegraphics[width=.98\linewidth]{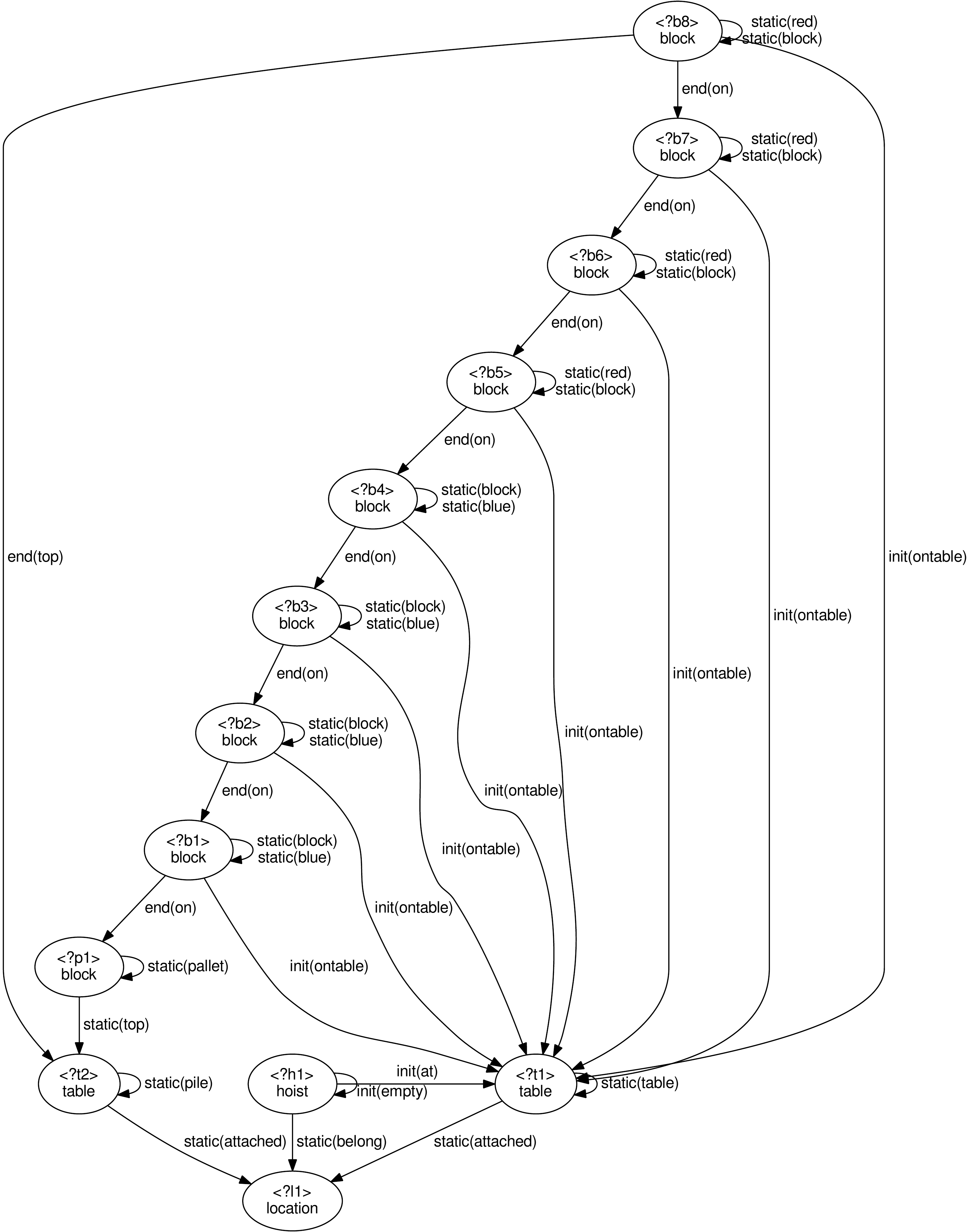}
        \vspace{-6pt}
        \caption{A $2$-valued structure $C$.}
        \label{fig:concrete_structure}
    \end{subfigure}
    \\\vspace{-2pt}
    \begin{subfigure}[b]{.66\linewidth}
        \includegraphics[width=\linewidth]{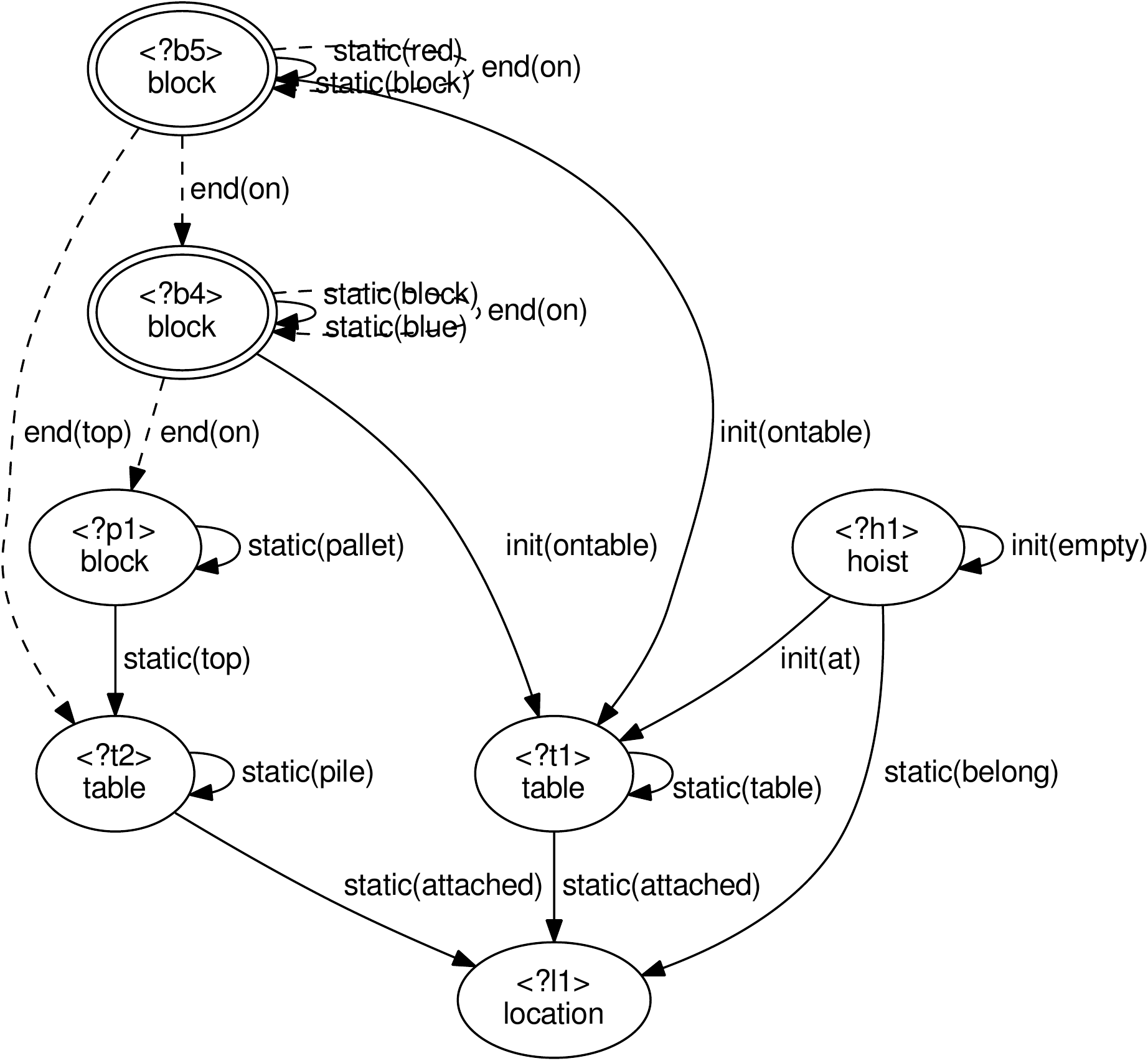}
        \vspace{-16pt}
        \caption{A $3$-valued structure $S=\beta(C)$.}
        \label{fig:abstract_structure}
    \end{subfigure}\vspace{-8pt}
    \caption{Canonical abstraction of the (generalized) `stack' experience 
    (see Listing~\ref{lst:experience}). 
    Nodes constitute the universe of a structure and edges represent the 
    truth-values of the key-properties. The nodes 
    {\fontfamily{cmss}\selectfont\small{?b4}} and 
    {\fontfamily{cmss}\selectfont\small{?b5}} are summary objects that represent 
    all blocks of the same properties (i.e., \CM{blue/red}) in an experience 
    (or in a problem).}
    \label{fig:abstraction}
\vspace{-15pt}
\end{figure}

\figref{abstraction}(a) shows a $2$-valued structure $C$ 
representing the (generalized) experience in Listing~\ref{lst:experience}. 
In this example, the universe, the set of predicates and truth-values
(interpretations) of the predicates over the universe of $C$ are as follows:
\vspace{-9pt}
{\fontfamily{cmss}\selectfont\small
\begin{align*}
U     &= \{ \text{?t1,?t2,?l1,?h1,?p1,?b1,?b2,?b3,}\dots \} \\
\Voc  &= \{ \text{(static(block)),(init(ontable)),(end(on)),} \dots \} \\
\iota &= \{ \text{(static(table ?t1)),(static(block ?b1)),} \dots \}\enspace.
\end{align*}}
\vspace{-10pt}

The scope inference (i.e., abstraction) is based on Kleene's $3$-valued logic 
\cite{kleene1952introduction}, which extends Boolean logic by introducing an 
indefinite value $\Half$, to denote either $0$ or $1$.

\begin{definition}\label{def:3_valued}
A \emph{$3$-valued logical structure}, also called an
\emph{abstract structure},  over a finite set of predicates $\Voc$ is a
pair $S=\langle U, \iota\rangle$ where $U$ is the universe of the
$3$-valued structure 
and $\iota$ is the interpretation function
mapping predicates to their truth-values in the structure: for every
predicate $p\in \Voc$ of arity $k$, $\iota(p):U^k \to \{0,1,\Half\}$.

A $3$-valued structure may include \emph{summary objects}, i.e., objects
that correspond to one or more objects in a $2$-valued structure represented
by the $3$-valued structure. 
\end{definition}

\figref{abstraction}(b) shows a $3$-valued structure $S$ 
of the $2$-valued structure $C$ in \figref{abstraction}(a). 
Double circles stand for summary objects and solid (dashed) arrows 
represent truth-values of $1$ ($\Half$). 
Intuitively, because of the summary objects, the $3$-valued structure $S$ 
represents the $2$-valued structure $C$ and all other `stack' problems that 
have exactly one \emph{table}, one \emph{pile}, one \emph{location}, one 
\emph{hoist}, one \emph{pallet}, and at least one \emph{blue block} and 
one \emph{red block} such that the blocks are initially on a table and 
finally red blocks are on top of blue blocks~in~a~pile. 

The objects in a $2$-valued structure are merged into a summary 
object in a $3$-valued structure as follows:

\begin{definition}
\label{def:canonical_name}
Let $\Voc^{(k)}$ denotes a set of predicates of arity $k$, 
and $C=\langle U, \iota\rangle$ is a $2$-valued structure. 
The \emph{canonical name} of an object $u \in U$, also called an 
\emph{abstraction predicate}, denoted by $\Can(u)$, is a set of 
unary predicates that hold for $u$ in $C$:
\(
\Can(u) = \{ p\in \Voc^{(1)} \mid \iota(p)(u)=1\}.
\)
\end{definition}

For example, the canonical names of the objects in structure $C$ of 
\figref{abstraction}(a) are the following:
\[\fontfamily{cmss}\selectfont\small
\begin{array}{rcl}
\Can(\text{?t1}) &=& \{\text{static(table)}\}\\
\Can(\text{?t2}) &=& \{\text{static(pile)}\}\\
\Can(\text{?l1}) &=& \{\text{static(location)}\}\\
\Can(\text{?h1}) &=& \{\text{static(hoist),init(empty)}\}\\
\Can(\text{?p1})   &=& \{\text{static(pallet)}\}\\
\Can(\text{?b1..?b4})   &=& \{\text{static(block),static(blue)}\}\\
\Can(\text{?b5..?b8})   &=& \{\text{static(block),static(red)}\}\enspace.
\end{array}
\]

\begin{definition}
\label{def:summary}
Let $\Voc$ be a set of predicates, $C=\langle U, \iota\rangle$ a $2$-valued 
structure, and $S=\langle U', \iota'\rangle$ a $3$-valued structure, over $\Voc$.
A \emph{summary object} $w \in U'$ corresponds to two objects $(u,v) \in U$, if 
$\Can(u)=\Can(v)$. 
\end{definition}

For example, the objects ({\fontfamily{cmss}\selectfont ?b1..?b4}) in 
\figref{abstraction}(a) with the same canonical name are 
merged into a summary object. 

Each $2$-valued structure $C$ is represented by its canonical abstraction, 
i.e., a $3$-valued structure in which all objects in $C$ with the same 
canonical name are merged into a summary element of that canonical name:

\begin{definition}
\label{def:CanonicalAbstraction}
Let $\Voc$ be a set of predicates, and $C=\langle U, \iota\rangle$ 
a $2$-valued structure. 
The \emph{Canonical abstraction} of $C$, denoted by $\beta(C)$, is
a $3$-valued structure $S=\langle U', \iota'\rangle$ as follow:
\begin{align*} 
&U' = \{ \Can(u) \mid u \in U\}\\
&\iota'(p)(t_1',\ldots,t_k') =\\
&\KJoin\limits_{t_1,\ldots,t_k} \{ \iota(p)(t_1,\ldots,t_k) \mid \forall i=1..k.\ t_i' = \Can(t_i) \} \enspace.
\end{align*} 

The canonical abstraction is based on Kleene's \emph{join operation} 
$\kjoin : 2^{\{0,1,\Half\}} \rightarrow \{0,1,\Half\}$, 
which overapproximates non-empty sets of logical values as follows:
\[\fontfamily{cmss}\selectfont
 \kjoin \Voc = \left\{
                         \begin{array}{ll}
                           v, & \hbox{if}\enspace\Voc=\{v\}\hbox{;} \\
                           \Half, & \hbox{otherwise.}
                         \end{array}
                       \right.
\]
\vspace{-2pt}
\end{definition}

Kleene's join operation determines the truth-value (interpretation) 
of key-properties in a $3$-valued structure. The interpretation of a 
key-property in the $3$-valued structure is $1$ (solid arrows) if that 
key-property exists for all objects of the same canonical name in the 
$2$-valued structure; the truth-value is $\Half$ if the key-property 
exists for some objects of the same canonical name (dashed arrows); 
and $0$ otherwise.

The inferred scope is finally represented 
as a set of key-properties. 
A summary object \sans{?o} is represented by 
a proposition of the form \sans{(summary ?o)}. 
An indefinite (i.e., $\Half$-valued) key-property $p$ appears as 
\sans{(maybe $p$)}. 
\lstref{schema_loop_prec} shows the inferred scope for the `stack' activity schema.

\begin{lstlisting}[style=customlst,
    label=lst:schema_loop_prec,
    float=t,
    captionpos=b,
    % numbersep=-4.5pt,
    %breaklines=true,
    % abovecaptionskip=10pt,
    morekeywords={parameters, domain, objects,
    define, activity, schema, method, abstract, plan, scope},
    caption={The scope of the activity schema for the `stack' task.
    }]
(:method stack
 :parameters    (?table ?pile)
 :scope         ((summary ?b4)
                 (summary ?b5)
                 (static(red ?b5))
                 (static(attached ?t2 ?l1)) 
                 (static(attached ?t1 ?l1))
                 (static(block ?b4)) 
                 (static(block ?b5))
                 (static(belong ?h1 ?l1))
                 (static(pallet ?p1))
                 (static(blue ?b4)) 
                 (static(pile ?t2))
                 (static(top ?p1 ?t2))
                 (static(table ?t1)) 
                 (init(ontable ?b4 ?t1))
                 (init(ontable ?b5 ?t1)) 
                 (init(at ?h1 ?t1))
                 (init(empty ?h1))
                 (maybe(end(top ?b5 ?t2)))
                 (maybe(end(on ?b5 ?b5))) 
                 (maybe(end(on ?b4 ?p1)))
                 (maybe(end(on ?b5 ?b4))) 
                 (maybe(end(on ?b4 ?b4))))
 :abstract-plan ($\text{the same as in \lstref{schema_loop}}$))
\end{lstlisting}

\section{Testing the Scope of Applicability}
\label{sec:tvla_execution}

An activity schema is applicable for solving a task problem~if the task 
problem is \emph{embedded} in the scope of the activity schema (i.e., 
the task problem maps onto the scope of the activity schema). 
For this purpose, we convert a task problem 
$\mcr{P}=\langle t,\sigma,s_0,g \rangle$ 
into a $2$-valued structure 
(as described in the previous section), and then test if the 
obtained $2$-valued structure is embedded in the scope of an activity schema:

\begin{definition}
\label{def:embedding}
We say that a $2$-valued structure (i.e., a task problem represented 
in a $2$-valued structure) $C=\langle U, \iota\rangle$ is \emph{embedded} 
in a $3$-valued structure (i.e., the scope of an activity schema) 
$S=\langle U', \iota'\rangle$, denoted by $C \sqsubseteq S$, if there exists 
a function $f:U \to U'$ such that $f$ is surjective and for every predicate 
$p$ of arity $k$ and tuple of objects $u_1,...,u_k \in U$, one of the following 
conditions holds:
\begin{equation}
\label{eq:embedding}
\begin{array}{c}
\iota(p)(u_1,...,u_k) = \iota'(p)(f(u_1),...,f(u_k))
\quad\text{or} \\
\iota'(p)(f(u_1),...,f(u_k)) = \Half \enspace.
\end{array}
\end{equation}

\noindent
Further, a $3$-valued structure $S$ represents the set of $2$-valued structures 
embedded in it: $\{C \mid C \sqsubseteq S\}$.
\end{definition}

\begin{proposition}
Canonical abstraction is sound  with respect to the embedding relation.
That is, $C \sqsubseteq \beta(C)$ holds for every $2$-valued structure $C$.
\end{proposition}

We implemented and integrated an 
\textsc{Embedding} function into the EBPDs' planning system which finds 
an applicable activity schema $m$ with the scope of applicability $S$ 
to a task problem $\mcr{P}$, by checking whether 
$\Struc(\mcr{P}) \sqsubseteq S$ holds. 


\section{Experimental Resutls}\label{sec:experiments}

We implemented a prototype of this system in \textsc{Prolog} and used 
TVLA \cite{lev2004tvla} as an engine, implemented in \textsc{Java}, 
for computing the scope of applicability of activity schemata. 
We develop two EBPDs based on classical planning domains and evaluate 
our system in classes of tasks in these domains. 

\medskip\noindent\textbf{\textsc{STACK}.}
In the first experiment, we develop the \StackDom domain, based 
on the blocks world domain, containing the concrete planning operators, 
\sans{move/4}, \sans{pick/4}, \sans{put/4}, \sans{stack/5}, \sans{unstack/5}, 
and the abstract planning operators, \sans{pick/3}, \sans{put/3}, 
\sans{stack/4}, \sans{unstack/4} (i.e., the numbers indicate arities). 
The main objective of this experiment is to learn different activity 
schemata (tasks) with the same goal but different scopes of applicability, 
and to evaluate how the scope testing (embedding) function allows the system 
to automatically find an applicable activity schema to a given task problem. 

In the paper, we described a class of `stack' problems with an experience 
(in Listing~\ref{lst:experience}), a learned activity schema 
(in Listing~\ref{lst:schema_loop}), and its scope of applicability (in 
Listing~\ref{lst:schema_loop_prec} and Fig.~\ref{fig:abstraction}(b)). 
Additionally, we define three other classes of the `stack' problems with the same 
goal but different initial configurations as follows: (\emph{i}) a pile of red and 
blue blocks, with red blocks at the bottom and blue blocks on the top; (\emph{ii}) 
a pile of alternating red and blue blocks, with a blue block at the bottom and a red 
block on the top; and (\emph{iii}) a pile of alternating red and blue blocks, with a 
red block at the bottom and a blue block on the top. 
In all classes of problems, the goal is to make a new pile of red and blue blocks 
with blue blocks at the bottom and red blocks on the top. 

\begin{figure}[!t]
    \centering
    \begin{subfigure}[!t]{\linewidth}
        \centering
        \captionsetup{width=\linewidth}%
        \includegraphics[width=.65\linewidth]{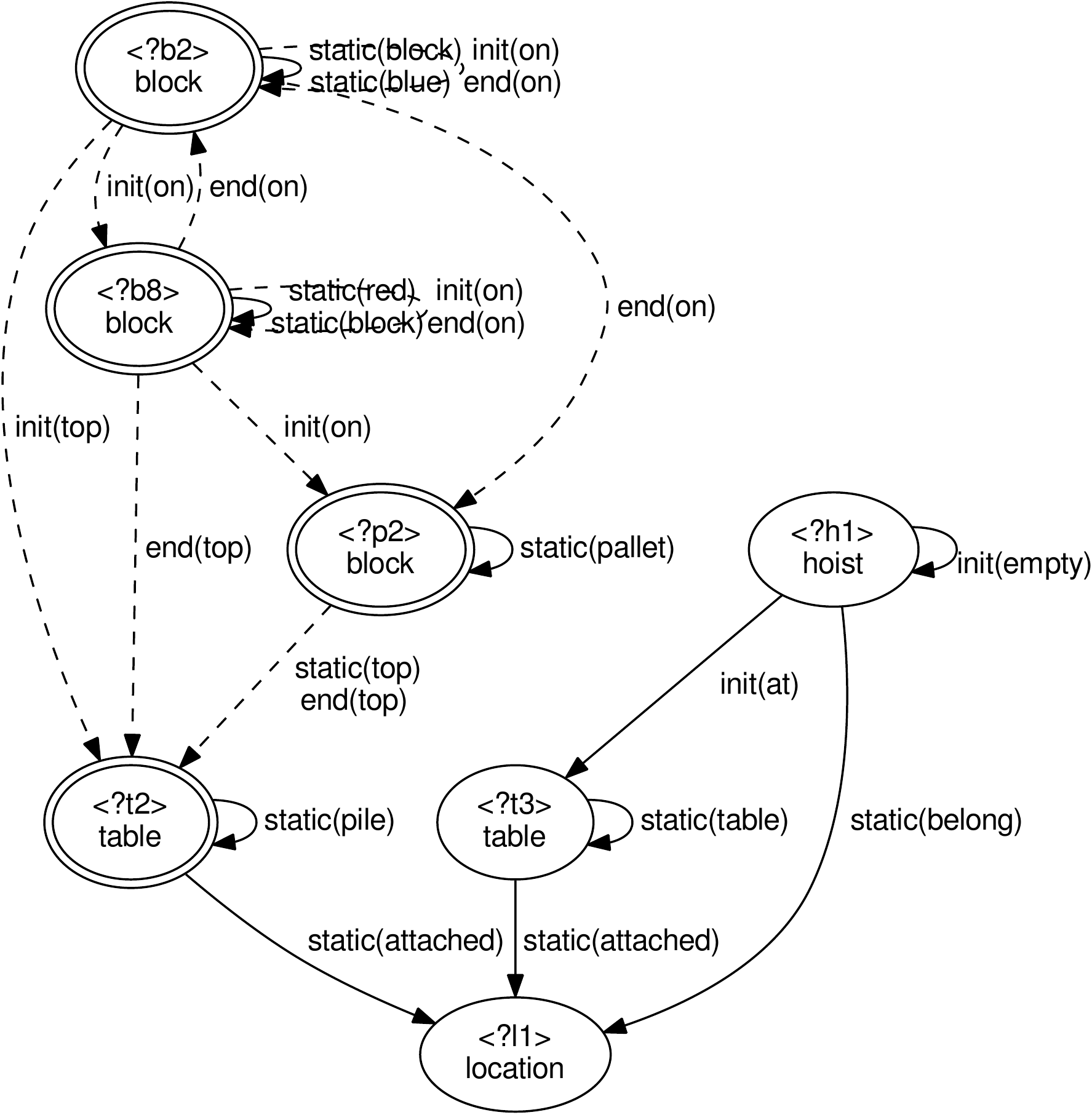}
        \caption{This scope of applicability (abstract structure) represents all 
        `stack' problems that have exactly one \emph{table} and at least one \emph{pile}, 
        one \emph{pallet}, one \emph{blue block} and one \emph{red block} such that 
        blue blocks are initially on top of red blocks and finally red blocks are 
        on top of blue blocks (on a pallet) on a pile.}
        \label{fig:abstract_stack3}
    \end{subfigure}
    \\\vspace{0pt}
    \begin{subfigure}[!t]{\linewidth}
        \centering
        \captionsetup{width=\linewidth}%
        \includegraphics[width=.795\linewidth]{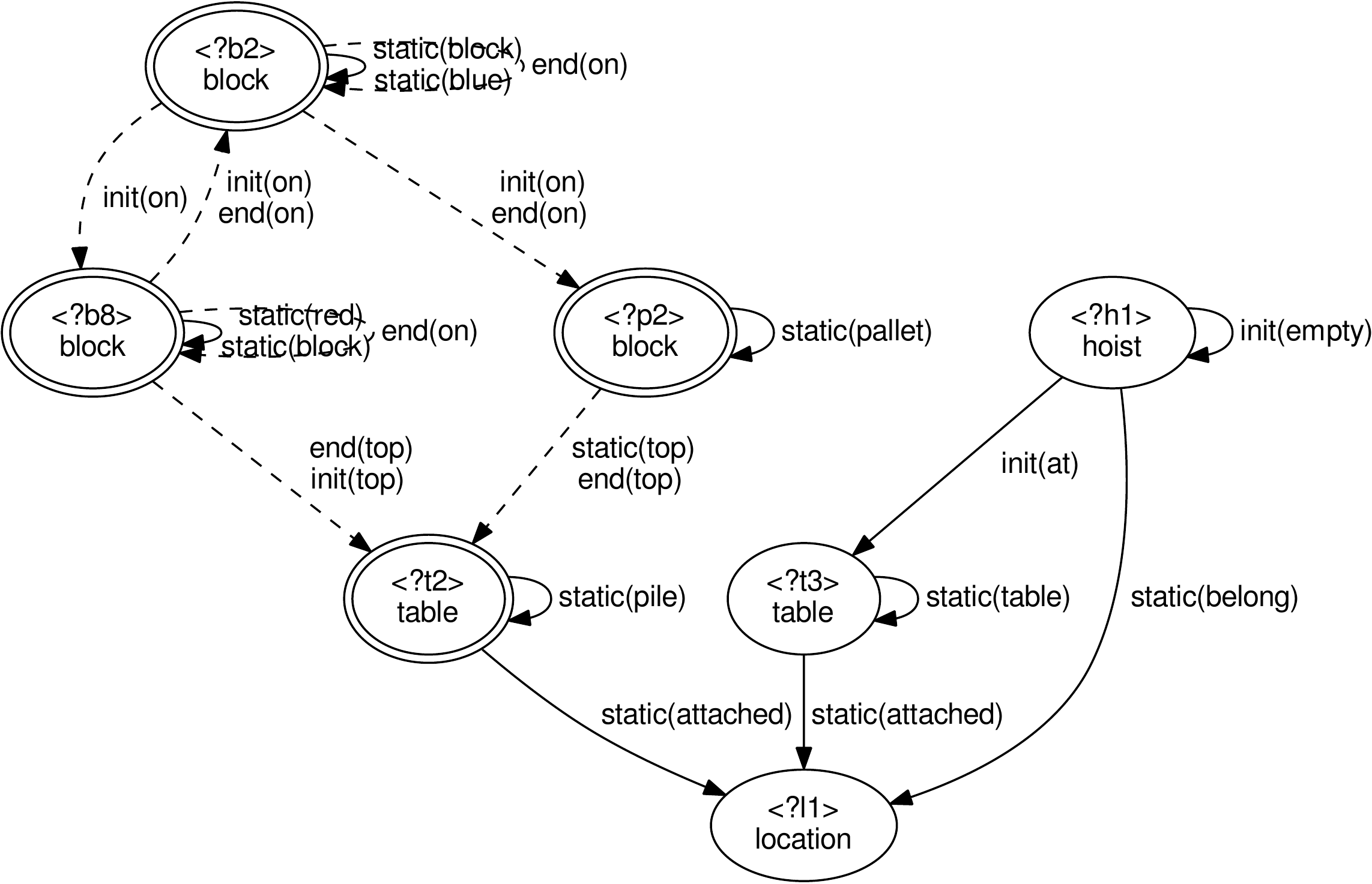}
        \caption{This scope of applicability represents all `stack' problems that 
        have exactly one \emph{table} and at least one \emph{pile}, one \emph{pallet}, 
        one \emph{blue block} and one \emph{red block} such that alternate red and blue 
        blocks are initially on a pile with a blue block at the bottom (on a pallet) 
        and a red block on top and finally red blocks are on top of blue blocks.}
        \label{fig:abstract_stack4}
    \end{subfigure}
    \\\vspace{0pt}
    \begin{subfigure}[!t]{\linewidth}
        \centering
        \captionsetup{width=\linewidth}%
        \includegraphics[width=.65\linewidth]{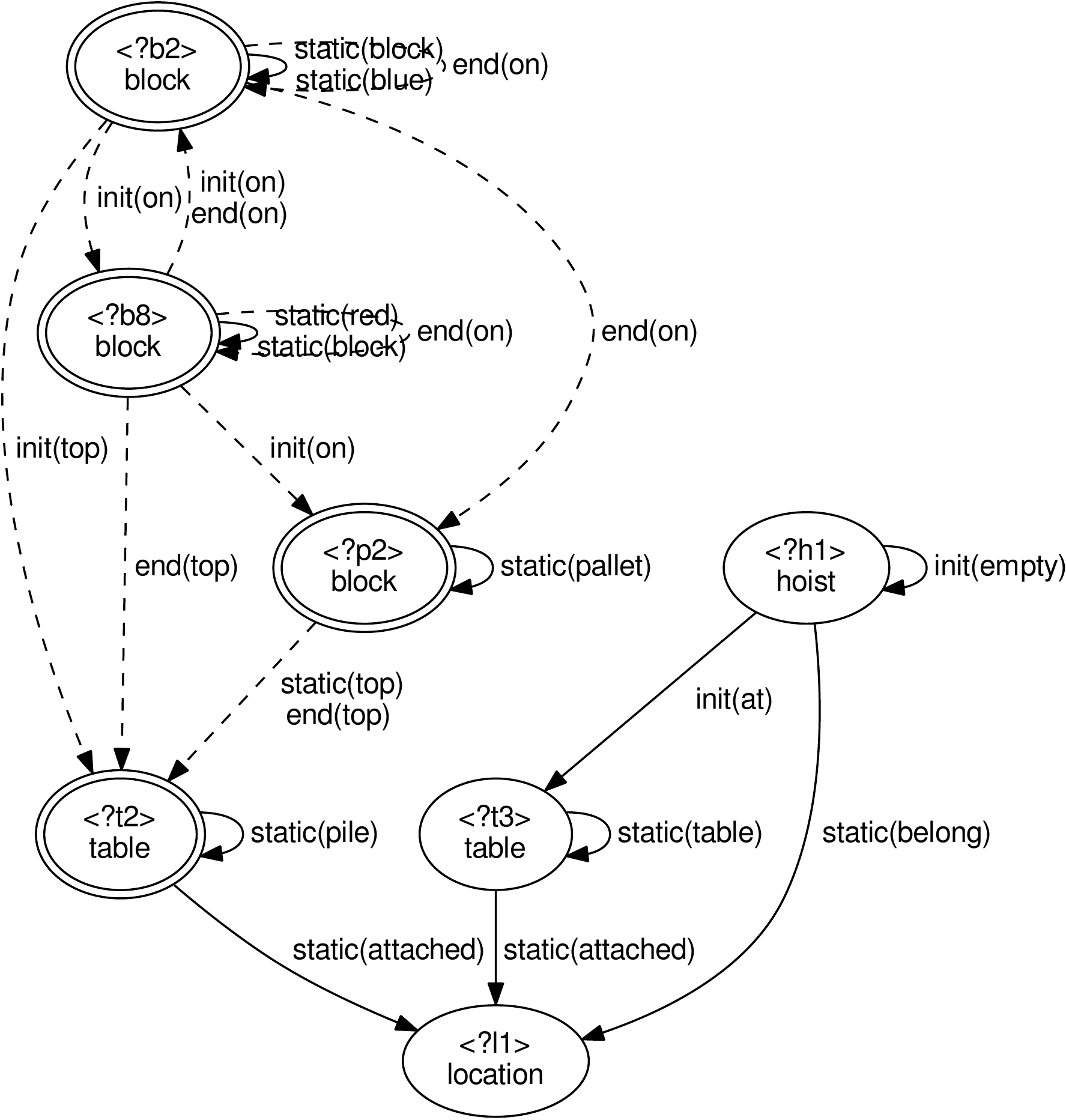}
        \caption{This scope of applicability represents all `stack' problems 
        that have exactly one \emph{table} and at least one \emph{pile}, 
        one \emph{pallet}, one \emph{blue block} and one \emph{red block} 
        such that alternate red and blue blocks are initially on a pile with 
        a red block at the bottom (on a pallet) and a blue block on top and 
        finally red blocks are on top of blue blocks.}
        \label{fig:abstract_stack5}
    \end{subfigure}
    \caption{The scope of applicability, i.e., canonical abstraction, of the 
    additional three classes of the `stack' task in the \StackDom domain.
    }
    \label{fig:stack_scopes}
\vspace{-35pt}
\end{figure}

To show the effectiveness of the proposed scope 
inference, we simulated an experience (containing an equal number of $20$ blocks 
of red and blue colors) in each of the above classes. Based on these experiences 
the system generates three activity schemata with distinct scopes of applicability 
(see Fig.~\ref{fig:stack_scopes}). 

\begin{figure}[!t]
  \centering
  \includegraphics[width=\linewidth]{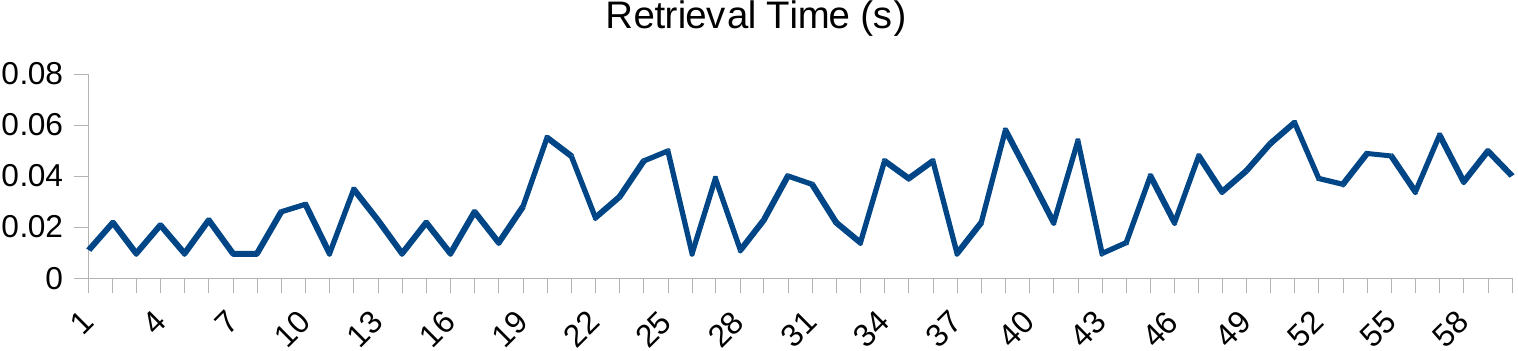}
  \vspace{-15pt}
  \caption{CPU time used by SBP to find an applicable activity 
  schema (among 4) for solving problems in the \StackDom domain.}
  \label{fig:stack_retrieval}
  \vspace{-5pt}
\end{figure}

\begin{figure}[!t]
  \centering
  \begin{subfigure}[b]{.498\linewidth}
    \includegraphics[width=\linewidth]{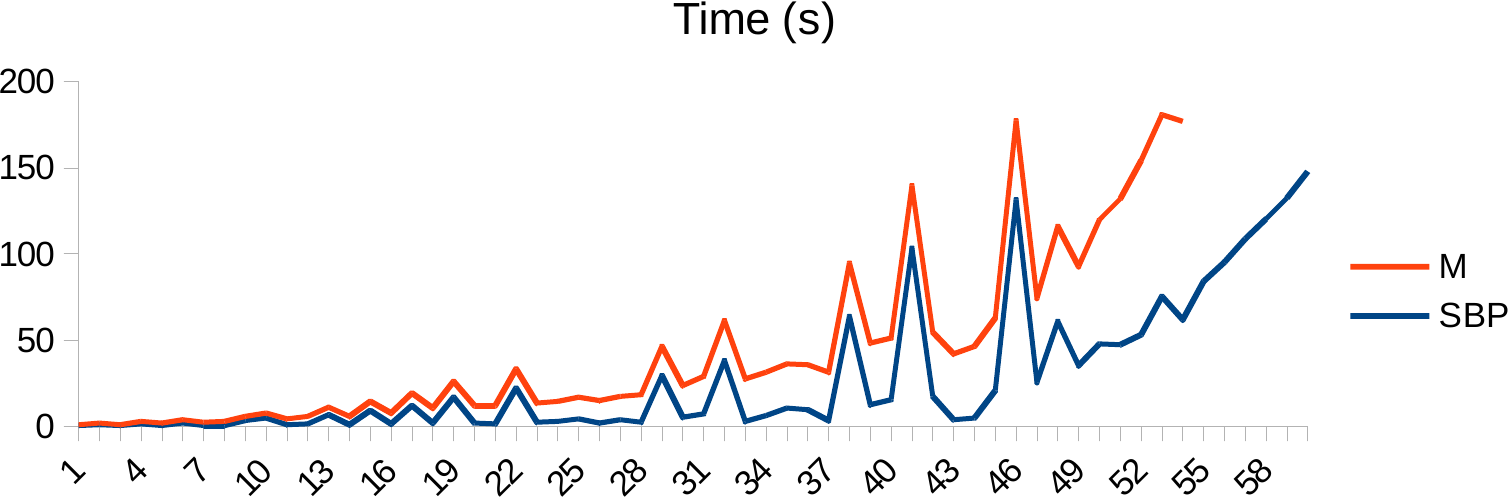}
  \end{subfigure}\hspace{-1.4pt}
  \begin{subfigure}[b]{.498\linewidth}
    \includegraphics[width=\linewidth]{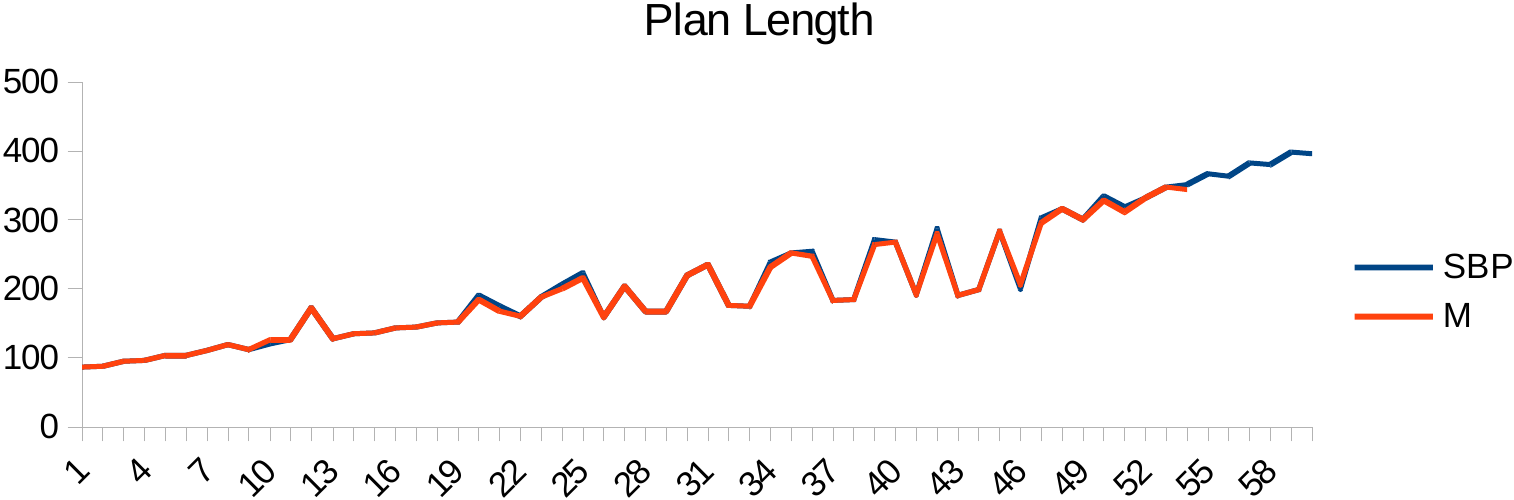}
  \end{subfigure}\\\vspace{4pt}
  \begin{subfigure}[b]{.498\linewidth}
    \includegraphics[width=\linewidth]{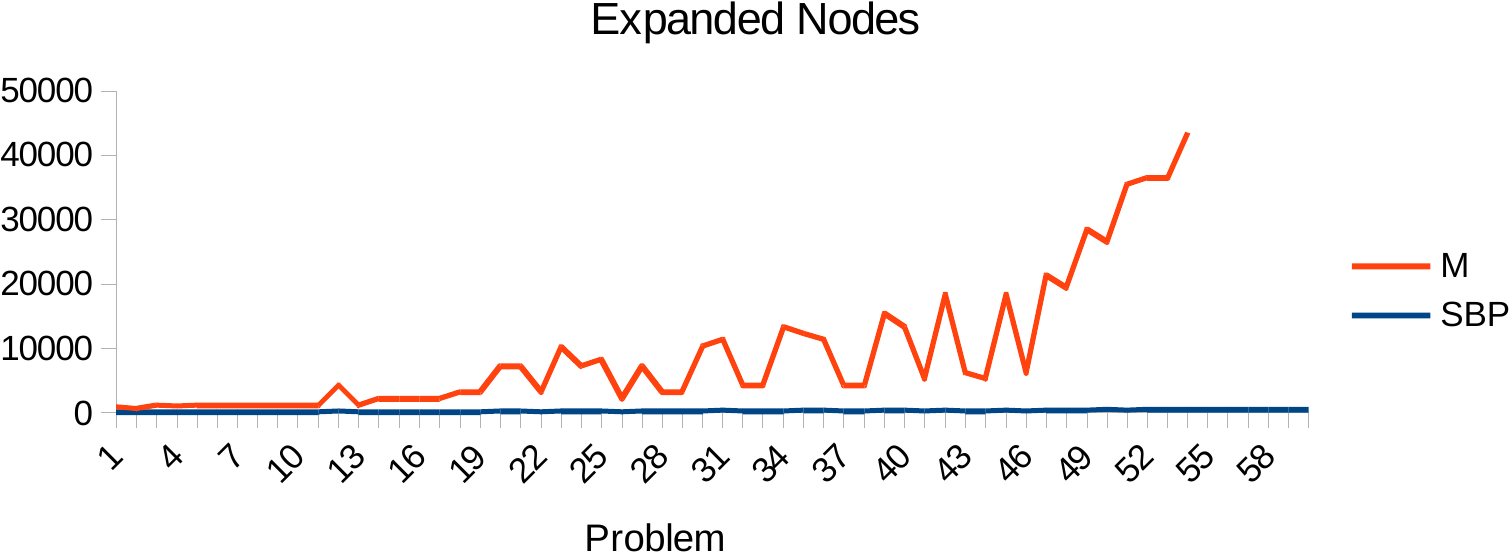}
  \end{subfigure}\hspace{-1.4pt}
  \begin{subfigure}[b]{.498\linewidth}
    \includegraphics[width=\linewidth]{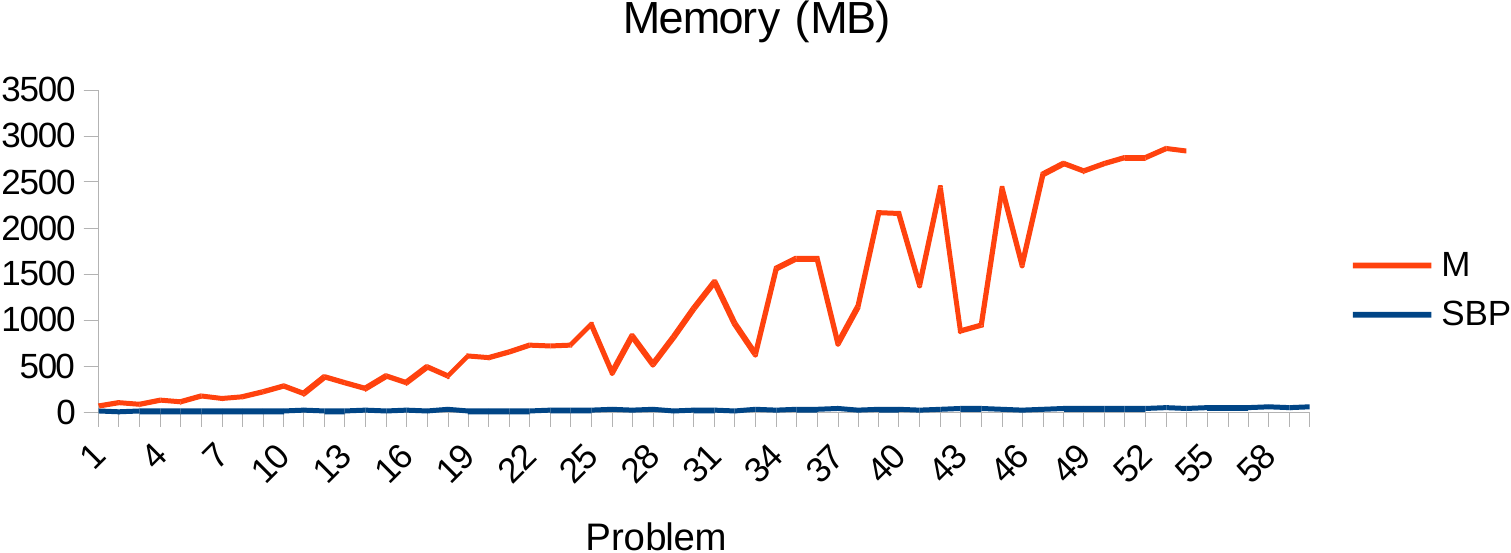}
  \end{subfigure}\\
  \vspace{-5pt}
  \caption{Performance of the SBP and \textsc{Madagascar} (M) 
  in the \StackDom domain.}
  \label{fig:stack_chart}
  \vspace{-5pt}
\end{figure}

To evaluate the system over the learned activity schemata, we randomly generated 
$60$ task problems in all four classes of the `stack' tasks, ranging from $20$ to 
$50$ equal number of red and blue blocks in each problem. 
In this experiment, the system found applicable activity schemata to solve given 
task problems in under $60\text{ms}$ for testing the scope of applicability 
(see Fig.~\ref{fig:stack_retrieval})
and then successfully solved all problems. 
To show the efficiency of the system, we also evaluated and compared the 
performance of the SBP with a state-of-the-art planner, \textsc{Madagascar}  
\cite{RINTANEN201245}, based on four measures: time, memory, number of evaluated 
nodes and plan length (see Fig.~\ref{fig:stack_chart}).
In this experiment, SBP was extremely efficient in terms of memory and evaluated 
nodes in the search tree.
Note that the time comparison is not accurate, since SBP has 
been implemented in \textsc{Prolog}, in contrast to \textsc{Madagascar} that has 
been implemented in \textsc{C++}. 

\begin{figure}[!t]
  \centering
  \begin{subfigure}[b]{.58\linewidth}
  \centering
        \includegraphics[width=\linewidth]{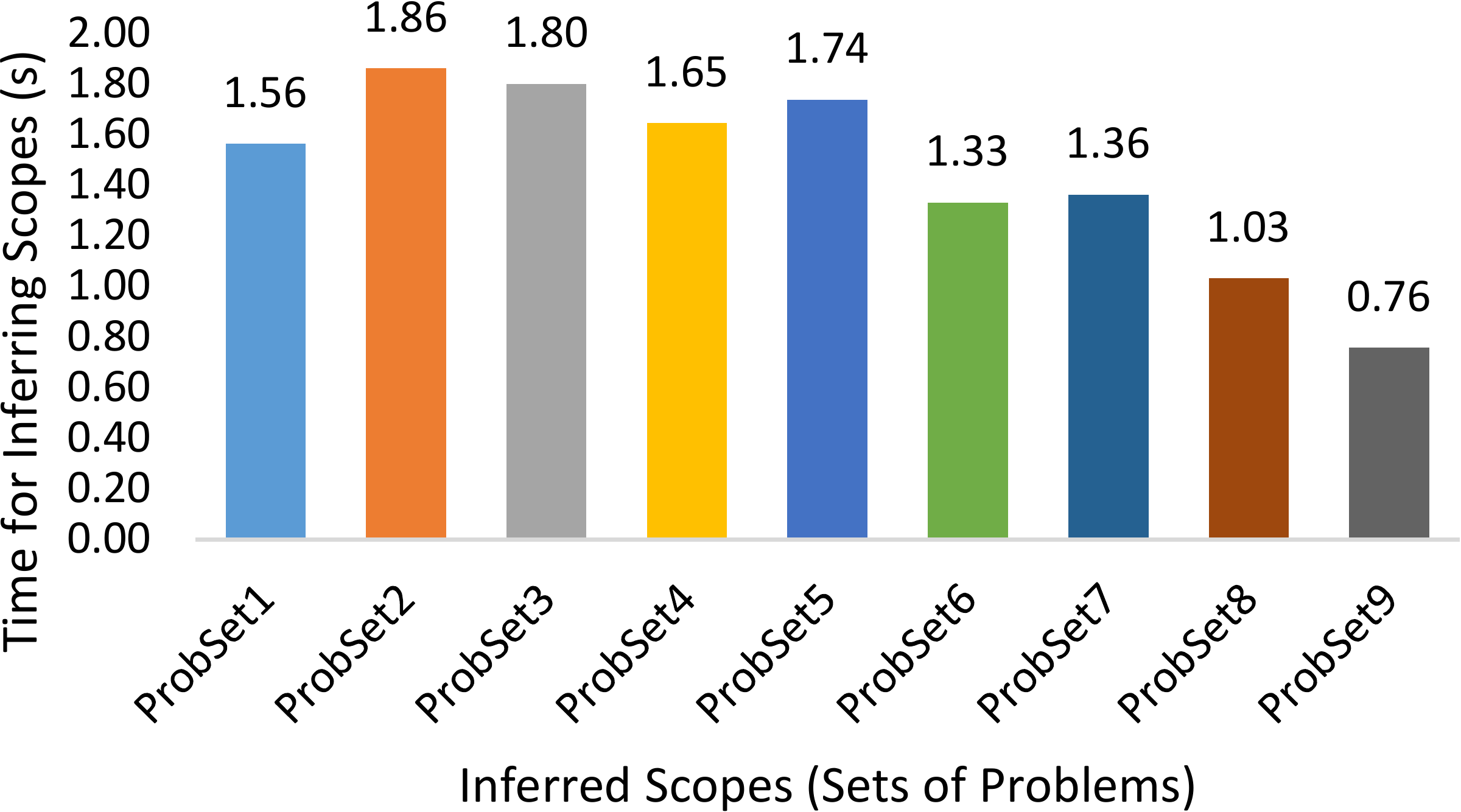}
        \vspace{-15pt}\caption{} 
        \label{fig:rover_scope}
  \end{subfigure}\hspace{-2.4pt}
  \begin{subfigure}[b]{.42\linewidth}
  \centering
        \includegraphics[width=\linewidth]{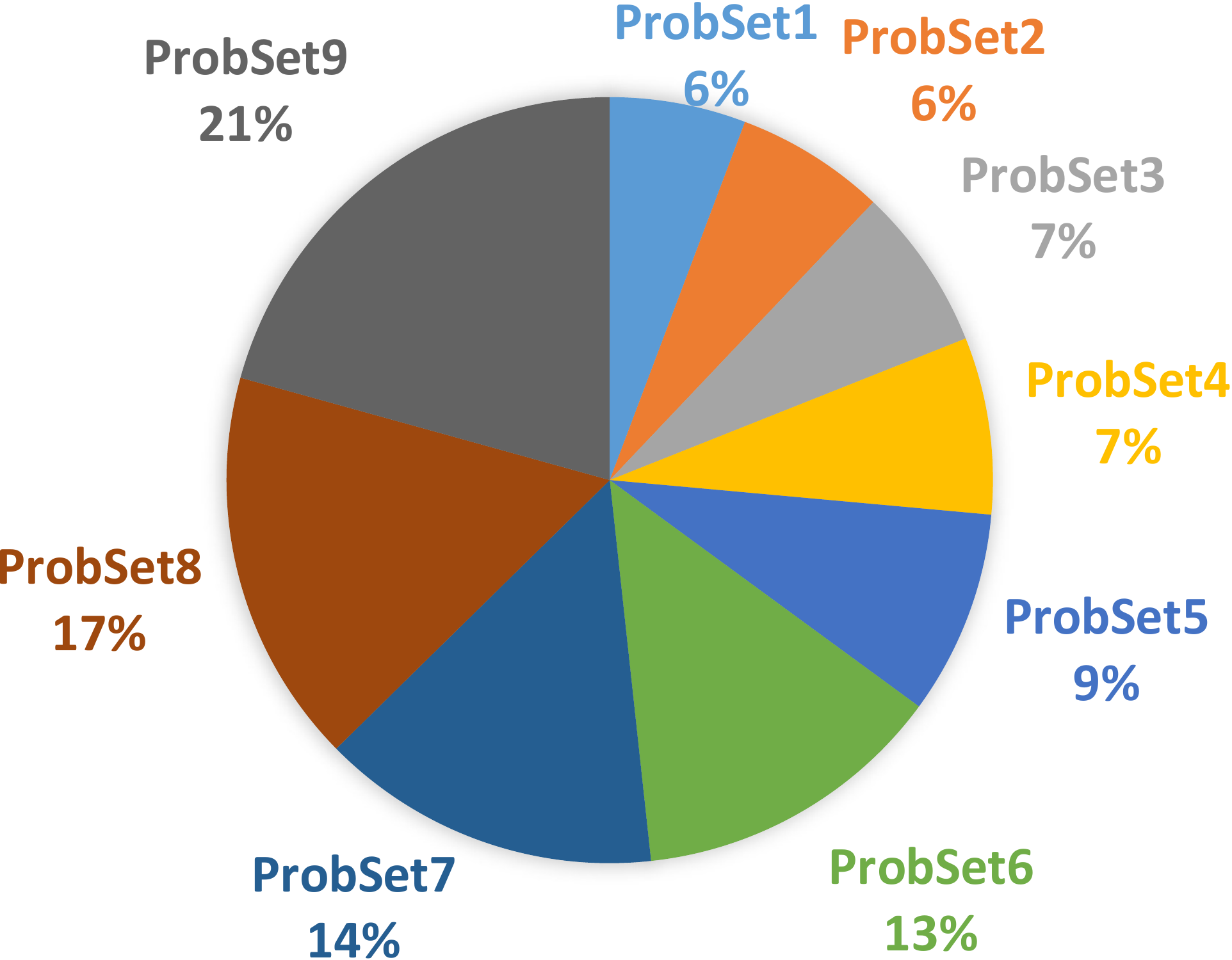}
        \vspace{-15pt}\caption{} 
        \label{fig:rover_portion}
  \end{subfigure}\\\vspace{-10pt}
  \caption{CPU time used by TVLA to classify the problems (a), 
  and distribution of the problems in the obtained problem sets 
  (b) in the \Rover domain}
  \label{fig:rover}
  \vspace{-7pt}
\end{figure}

\begin{figure}[!t]
  \centering
  \includegraphics[width=\linewidth]{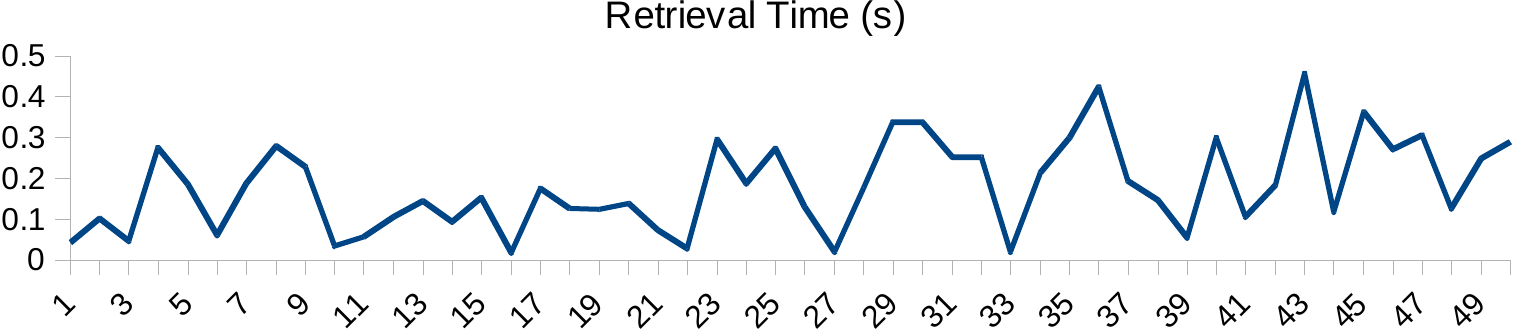}
  \vspace{-15pt}
  \caption{CPU time used by SBP to find an applicable activity 
  schema (among 9) for solving problems in the \Rover domain.}
  \label{fig:rover_retrieval}
  \vspace{-5pt}
\end{figure}

\medskip\noindent\textbf{\textsc{ROVER}.}
In the second experiment, we used the \Rover domain from the 
3rd International Planning Competition (IPC-3). 
In this experiment, we adopt a different approach for evaluating 
the proposed scope inference technique. 
We randomly generated $50$ problems containing exactly $1$ rover 
and ranging from $1$ to $3$ waypoints, $5$ to $30$ objectives, $5$ 
to $10$ cameras and $5$ to $20$ goals in each problem. 
Using the scope inference procedure, the problems are classified 
into $9$ sets of problems. That is, problems that converge to the 
same $3$-valued structure are put together in the same set. 
Hence, each set of problems is identified with a distinct scope 
of applicability. 
Fig.~\ref{fig:rover_scope} shows the time required to classify 
the problems into different sets, i.e., the time required by TVLA 
to generate $3$-valued structures for the problems and test 
which problems converge to the same $3$-valued structure. 
Fig.~\ref{fig:rover_portion} shows the distribution of the 
problems in the obtained sets of problems. 
In each set of problems, we simulated an experience and generated 
an activity schema for problem solving. 
Fig.~\ref{fig:rover_retrieval} shows the time required to retrieve 
an applicable activity schema (among 9 activity schemata in this 
experiment) for solving given problems, i.e., the time required to 
check whether a given problem is embedded in the scope of an activity 
schema. 
SBP successfully solved all problems in each class. 
\footnote{\fontsize{8pt}{8pt}\selectfont
The original experiences, activity schemata and task problems 
in our experiments 
are available at: \url{http://bit.ly/2IwJFCu}.
}

\section{Conclusion and Future Work}\label{sec:conclusion}

Using TVLA we generated a set of conditions that determine the 
scope of applicability of an activity schema in experience-based planning 
domains (EBPDs). 
The inferred scope allows an EBPD system to automatically find an applicable 
activity schema for solving a task problem.
We validated this work in two classical planning domains. 
The initial results show good scalability, however, engineering optimizations 
are possible on the prototype implementation of the proposed algorithms. 
This work is extensively presented in \cite{mokhtari2019learning}.

\bibliographystyle{named}
\bibliography{ref}

\end{document}